\documentclass{bmvc2k}

\usepackage{color}
\usepackage{changes}
\usepackage{booktabs}
\usepackage{makecell}
\usepackage{wrapfig}
\usepackage{twemojis}
\usepackage{marvosym}

\usepackage{float}
\newfloat{figtab}{htb}{fgtb}
\makeatletter
  \newcommand\figcaption{\def\@captype{figure}\caption}
  \newcommand\tabcaption{\def\@captype{table}\caption}
\makeatother

\usepackage{amssymb}

\title{High-Fidelity Eye Animatable Neural Radiance Fields for Human Face}

\addauthor{Hengfei Wang}{hxw080@student.bham.ac.uk}{}
\addauthor{Zhongqun Zhang}{zxz064@student.bham.ac.uk}{}
\addauthor{Yihua Cheng\textsuperscript{\Letter}}{y.cheng.2@bham.ac.uk}{}
\addauthor{Hyung Jin Chang}{h.j.chang@bham.ac.uk}{}

\addinstitution{
 School of Computer Science\\
 University of Birmingham\\
 Birmingham, UK
}

\runninghead{Hengfei Wang, Zhongqun Zhang, Yihua Cheng, Hyung Jin Chang}{DeNeRF}

\def\etal{\emph{et al}\bmvaOneDot}

\begin{document}

\maketitle

\begin{abstract}

Face rendering using neural radiance fields (NeRF) is a rapidly developing research area in computer vision. While recent methods primarily focus on controlling facial attributes such as identity and expression, they often overlook the crucial aspect of modeling eyeball rotation, which holds importance for various downstream tasks. In this paper, we aim to learn a face NeRF model that is sensitive to eye movements from multi-view images. We address two key challenges in eye-aware face NeRF learning: \textit{how to effectively capture eyeball rotation for training} and \textit{how to construct a manifold for representing eyeball rotation}. To accomplish this, we first fit FLAME, a well-established parametric face model, to the multi-view images considering multi-view consistency. Subsequently, we introduce a new Dynamic Eye-aware NeRF (DeNeRF). 
DeNeRF transforms 3D points from different views into a canonical space to learn a unified face NeRF model.
We design an eye deformation field for the transformation, including rigid transformation, e.g., eyeball rotation, and non-rigid transformation.
Through experiments conducted on the ETH-XGaze dataset, we demonstrate that our model is capable of generating high-fidelity images with accurate eyeball rotation and non-rigid periocular deformation, even under novel viewing angles. Furthermore, we show that utilizing the rendered images can effectively enhance gaze estimation performance.
\end{abstract}

\section{Introduction}

\begin{figure*}[t]
\centering
\includegraphics[width=1.0\linewidth]{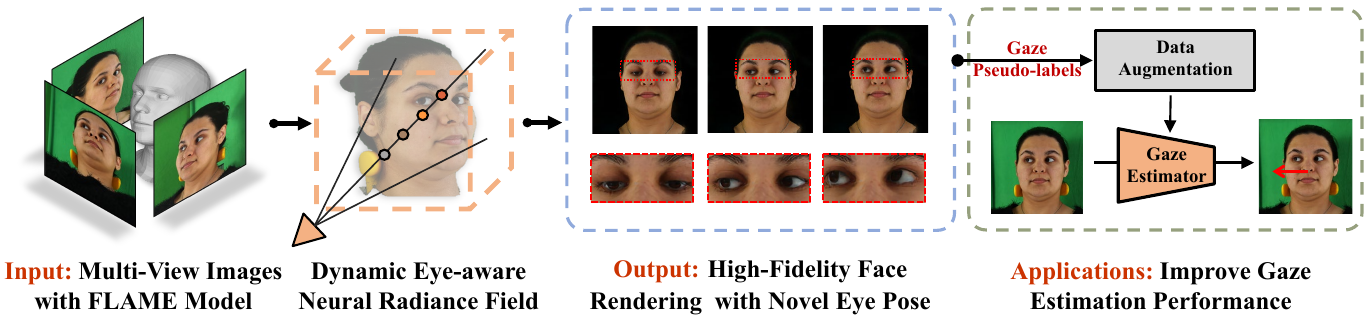}
\vspace{-0.5cm}
\caption{
The Dynamic Eye-aware Neural Radiance Field (DeNeRF) is designed to render high-fidelity faces with animatable eyes using a set of multi-view images. 
It enables face rending under novel view and eye pose.
By leveraging DeNeRF, we are able to obtain pseudo gaze labels from the model, which can be utilized to enhance gaze estimation methods.}
\label{fig:task}
\end{figure*}

Face rendering is an important task in computer vision and computer graphics.
It is widely demanded by applications such as virtual reality \cite{pai2016gazesim, mania2021gaze, henrysson2005face}, digital human \cite{demirel2007applications, jack2015human, leyvand2006digital} and CG film-making \cite{yang2022recursive, vanezis1989application, blanz2004exchanging}. Conventional methods fit a parametric face model based on a face template mesh \cite{li2017learning, feng2021learning}. 
Generative adversarial networks (GAN) directly render photo-realistic images with deep neural networks \cite{goodfellow2020generative, antipov2017face, gauthier2014conditional}.
Recent research has incorporated the Neural Radiance Fields (NeRF) \cite{mildenhall2021nerf} for face rendering. NeRF models encode 3D geometry and exhibit great multi-view consistency that enables face rendering under novel viewpoints. 
Additionally, various previous works explore parametric face NeRF models~\cite{hong2022headnerf}, which enable the control of facial attributes in rendered images such as identity and expression. 

While most face NeRF models focus on facial semantic attributes, they tend to overlook the importance of controllable eye movement in face rendering.
Eye movement can enhance image realism and is critical for multiple downstream tasks.
However, the eyeball is located inside the face and incompletely visible in face images. It is difficult to determine the precise 3D eyeball rotation and position  based on only face images.
Furthermore, the eyeball rotation is related to geometric architecture, which means the model should be rotation-aware. 
Eyeball rotation also leads to non-rigid deformation such as the periocular deformation.
Previous methods perform the rotation in feature space with gaze directions \cite{ruzzi2022gazenerf, zheng2020self}.
However, it is also non-trivial to obtain accurate gaze directions from images and such approaches often result in low-quality rendered images.

In this paper, we propose a novel approach called Dynamic Eye-aware NeRF (DeNeRF).
DeNeRF learns a dynamic face NeRF model from multi-view images, enabling face rendering with unseen eyeball and head poses.
We address two challenges in eye-aware NeRF learning: \textit{effectively capturing eyeball rotation} and \textit{constructing a suitable manifold for representing such rotation}.
To capture eyeball rotation, we begin by fitting a well-established parametric face model, FLAME \cite{li2017learning}, to the multi-view face images. FLAME is originally designed for face tracking from a single face image, and we modify the fitting process to account for consistency across multi-view images. This allows us to obtain parameters such as eyeball and head pose, which we then use for DeNeRF learning.

We further define a unified canonical space in DeNeRF.
Given a pixel in the observed images, we sample 3D positions based on view directions in the observation space.
We transform the 3D positions from the observation space into the canonical space with given poses.
We design an eye deformation field for the transformation, including rigid transformation ( \textit{e.g.}, eyeball rotation and head rotation) and non-rigid transformation.
We input the 3D positions in the canonical space into a NeRF model and render the pixel for alignment. 
To reduce the computational costs, we adopt a patch-based sampling approach \cite{schwarz2020graf}. We sample patches in images for alignment in each iteration.
To enforce realistic eye region, we generate an eye mask for each image and enlarge the sampling ratio in the eye region.

Overall, our contributions are three-fold: 
\vspace{-0.2cm}
\begin{itemize}
    \item We propose DeNeRF which learns a dynamic face NeRF model from multi-view images. 
To capture the eyeball pose accurately, we design a new fitting process for the FLAME model, ensuring consistency across multiple views.
\vspace{-0.2cm}
    \item We define a unified canonical space to construct a rotation-aware manifold. 
We transform 3D positions in the observation space into a canonical space based on an eye deformation field, including both rigid and non-rigid transformations. The DeNeRF is learned in the canonical space.
\vspace{-0.2cm}
    \item 
    DeNeRF enables high-fidelity face rendering under novel eyeball poses and head poses. Extensive experiments prove the rendered images can effectively enhance the performance of the downstream gaze estimation task.
\end{itemize}

\section{Related Work}

\textbf{Neural Radiance Field.} 
NeRF \cite{mildenhall2021nerf} proposed to learn implicit neural representations of a static scene from multi-view images, showing high-quality novel view synthesis. 
It models the continuous radiance field of a static scene by utilizing a mapping function that takes both a 3D spatial point $\mathbf{x}$ and view direction $\mathbf{d}$ as input, and outputs the corresponding RGB color $\mathbf{c}$ and volume density $\sigma$ values. A standard NeRF is parameterized with a Multi-Layer Perceptron (MLP) as 
\begin{equation}
    H_\theta:(\gamma(\mathbf{x}), \gamma(\mathbf{d})) \rightarrow(\mathbf{c}, \sigma),
\label{equ:nerf}
\end{equation}
where $\theta$ represents the parameters of the network and $\gamma$ refers to a positional encoding function \cite{mildenhall2021nerf, vaswani2017attention} that transforms $\mathbf{x}$ and $\mathbf{d}$ into a high-dimensional space.
Different from conventional generative models, NeRF is a 3D-aware model and represents 3D object/scene via implicit neural representations.
Novel views are generated from the implicit neural representation with volume rendering.
Further explorations~\cite{park2021nerfies, pumarola2021d, gafni2021dynamic, hong2022headnerf, peng2021animatable} adapt NeRF to represent dynamic scenes. Hong \etal~\cite{hong2022headnerf} acquire the latent codes of disentangled facial attributes from 3D morphable model.
They control the code to render images with different poses and identities.
Some methods also use NeRF for talking face generation \cite{guo2021ad,ye2023geneface,du2023dae}. They usually overlook the controllable eye movement in face rendering.

\textbf{Eye Image Synthesis.}
The precise manipulation of realistic eye imagery has proven essential across multiple domains of application.
Qin \etal \cite{qin2022learning} reconstruct 3DMM face model from multi-view images.
They can rotate a virtual camera to synthesize images in different camera poses but cannot rotate the eyeball.
Wood \etal \cite{Wood_2016_etra} build a virtual 3D morphable eye model with computer graphic algorithms. They can synthesize eye images with arbitrary gaze directions\cite{Wood_2016_etra, Wood_2016_ECCV, Wood_2015_ICCV}.
However, their synthetic images are usually not realistic enough.
Recently, generative model shows great potential in image generation.
Compared to 3D morphable model, generative model can generate more realistic images. 
Some methods use generative model to generate realistic eye images \cite{Shrivastava_2017_CVPR,He_2019_ICCV,Park_2019_ICCV,Yu_2019_CVPR}.
He \etal \cite{He_2019_ICCV} utilize GAN for gaze redirection task. They can synthesize large-scale gaze data by performing gaze redirection task on one eye image.
Ruzzi \etal \cite{ruzzi2022gazenerf} uses NeRF for gaze redirection tasks.
They train a NeRF model based on given gaze directions and perform rotation in feature space, which results in low-quality rendered images.

\section{Methodology}

\subsection{Multi-view Face Tracking}
\label{face tracking}

We first perform multi-view face tracking to acquire eyeball poses for training.
We use the well-known parametric face model FLAME \cite{li2017learning} for face tracking.
FLAME model fits facial parameters (such as pose and expression) from a single face image.
We modify the fitting process for the multi-view images
by projecting the fitted face model into multiple views and adding consistency loss for all views.

In practice, we input multi-view images and corresponding camera poses to the face tracker. 
We detect face landmarks, pupil centers, and face masks in each image~\cite{bulat2017far, yu2018bisenet}, where face masks are used to remove the background in the images.
We initialize the textured FLAME face model with zero parameters and project the face model to all views.
We render face images and perform pixel-wise alignment in the face appearance.
Note that, face masks are used to ensure the alignment is performed in the face region.
We can also obtain face landmarks and pupil centers from the FLAME model. 
We align them with the detected results for structure consistency.
We use $L_1$ loss for all alignments where we denote them as face appearance loss $\mathcal{L}_{appear}$, facial landmark loss $\mathcal{L}_{face}$ and  pupil center loss $\mathcal{L}_{pupil}$.
Overall, we optimize the multi-view face tracker by minimizing
\begin{equation}
\mathcal{L}_{tracking} = \left(\sum_{i=1}^{N}\left(\alpha\mathcal{L}_{pupil}^i + \beta\mathcal{L}_{face}^i + \gamma\mathcal{L}_{appear}^i\right)\right) / N,
\end{equation}
where $\alpha, \beta, \gamma$ are hyper-parameters, and we empirically set them as 10, 5, and 30. $N$ is the number of views which is 13 in our experiment. 
We show the details of the loss function and fitting result in the supplementary material. 

After fitting the FLAME model, we obtain facial parameters as the outcome. For DeNeRF Learning, we specifically choose four poses, namely two for the eyeballs, one for the jaw, and one for the neck. These poses are denoted as $p_i = \{R_i, t_i\}$, where $i$ ranges from 1 to 4. Collectively, we refer to all four poses as $\boldsymbol{p}$.

\subsection{Dynamic Eye-aware Neural Radiance Field}
\label{dynamic gaze nerf}

\begin{figure}[t]
\centering
\includegraphics[width=1.0\linewidth]{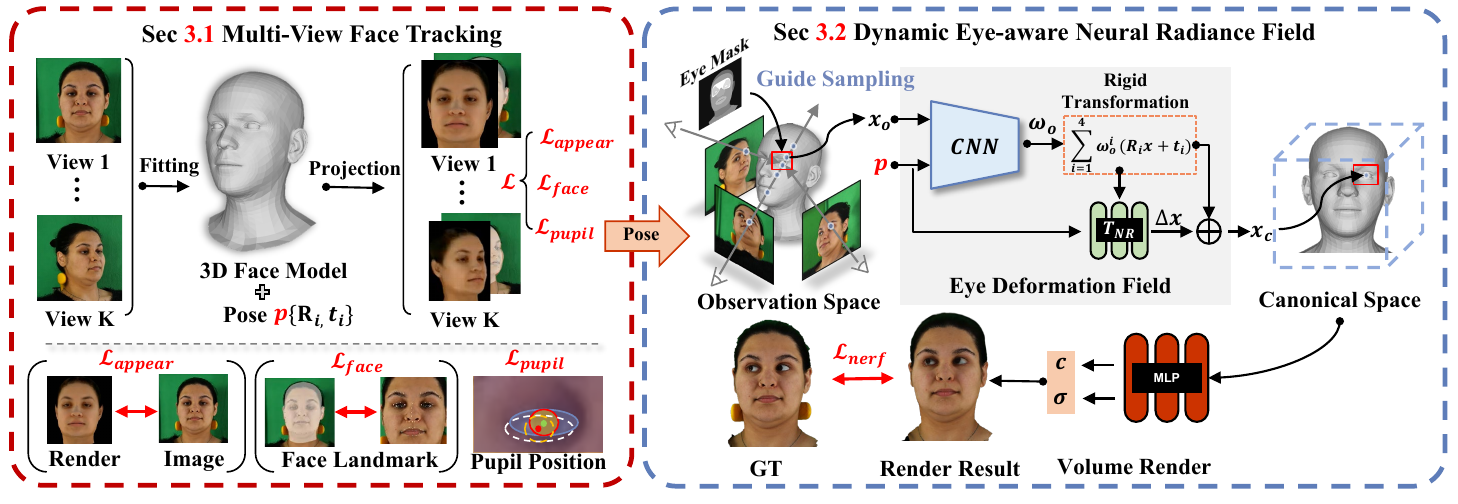}
\vspace{-0.5cm}
\caption{
Overview of DeNeRF model which specializes in creating high-fidelity face images that are aware of the eyes. It achieves this by training on multi-view images, and leveraging parametric model-based face tracking to establish multi-view consistency. From this data, DeNeRF fits a face model that includes poses for the eyeballs, jaw, and neck.
To facilitate the learning process, DeNeRF defines a canonical space and learns a unified model within it. Additionally, it introduces an eye deformation field that can transform points from observation space into the canonical space. This field is composed of both rigid and non-rigid transformation modules, enabling it to handle the complexity of eye movement and deformation.
}
\label{fig:pipeline}
\end{figure}

The input of DeNeRF contains multi-view images and poses $\boldsymbol{p}$ as well as camera poses.
Conventional NeRF-based methods implicitly learn a static geometric model. Although they can render images under novel views, they cannot control the content in the 3D model.

Our key idea is to learn a unified face NeRF model in the canonical space. 
Given an eyeball model and its pose, it is easy to rotate the eyeball from the observation space into the canonical space, and we can learn a unified eyeball model in the canonical space.
We perform the similar operation in DeNeRF.
We newly design an \textit{eye deformation field} on both rigid and non-rigid transformations. 
\textit{The eye deformation field} allows us to transform a point from the observation space into the canonical space.
We obtain the color and density of the point in the canonical space based on Eq.~(\ref{equ:nerf}) and learn a unified NeRF model in the canonical space.
 
\textbf{Eye Deformation Field.} 
We aim to learn a deformation field which transforms a point $\mathbf{x}_o$ in the observation space into the canonical space.
Intuitively, we would first rotate the 3D point based on the head pose, and then rotate it based on the eyeball pose if this point is in the eye region.
Unfortunately, this approach is not feasible due to the absence of an explicit mesh model. To address this issue, we propose a solution that involves applying various transformations, including head rotation and eyeball rotation, and combining them using learnable weights \cite{lewis2000pose}:
\begin{equation}
T_{\text {rigid}}\left(\mathbf{x}_o, p\right)=\sum_{i=1}^4 w_o^i(\mathbf{x}_o)\left(R_i \mathbf{x}_o+t_i\right).
\end{equation}
Note that we first convert all poses ${R_i, t_i}$ into a common coordinate system to facilitate the summation of transformed results.
Instead of directly estimating $w_o^i$, we learn $w_c^i$ as described in \cite{weng2022humannerf}, from which $w_o^i$ can be computed:
\begin{equation}
w_o^i(\mathbf{x}_o)=\frac{w_c^i \left(R_i \mathbf{x}_o+t_i\right)}{\sum_{k=1}^4 w_c^k \left(R_k \mathbf{x}_o+t_k\right)}.
\end{equation}
This process stabilizes the model and decreases the probability of collapse. 

$T_{rigid}$ exclusively deals with rigid transformations. However, facial appearance involves non-rigid transformations, such as periocular deformation. To address this, we introduce a non-rigid transformation field, denoted as $T_{NR}$, which is implemented as an MLP. This MLP generates an offset $\Delta x$ to augment the output of $T_{rigid}$. Additionally, we incorporate pose information $\boldsymbol{p}$ as supplementary inputs to the MLP. Consequently, the complete eye deformation field can be expressed as follows:
\begin{equation}
\mathbf{x}_c = \mathbf{\hat{x}} + T_{NR}(\mathbf{\hat{x}}, \boldsymbol{p}),
\end{equation}
where $\mathbf{\hat{x}} = T_{rigid}(\mathbf{x}_o, \boldsymbol{p})$ and $\mathbf{x}_c$ represents the position in the canonical space. Finally, we input the position $\mathbf{x}_c$ into a vanilla NeRF to obtain the color and density as Eq.~(\ref{equ:nerf}).

\subsection{Training}
We train DeNeRF to render $512\times512$ pixels images.
DeNeRF is trained with a photometric reconstruction loss in an end-to-end manner. 
To handle the high computational costs, we adopt a patch-based sampling approach \cite{schwarz2020graf}.
We randomly sample six patches with size $32\times32$ pixels from input images and 128 points for each ray.
We use LPIPS loss \cite{zhang2018unreasonable} ($\mathcal{L}_{\text {LPIPS }}$) and MSE loss ($\mathcal{L}_{\text {MSE}}$) for training as
\begin{equation}
\mathcal{L}_{\text {DeNeRF }}=\mathcal{L}_{\text {LPIPS }}(\mathbf{P}, \hat{\mathbf{P}})+\lambda{\mathcal{L}_{\text {MSE}}(\mathbf{P}, \hat{\mathbf{P}})},
\end{equation}
where $\mathbf{P}$ is the image patch and $\hat{\mathbf{P}}$ is the ground truth. We set $\lambda$ = 0.2 in the training.

\textbf{Eye-mask Guided Sampling.}
To address the issue of the eye region being too small compared to the rest of the face, we introduce an eye mask \cite{yu2018bisenet} to guide the ray sampling in DeNeRF. Specifically, we assign a larger sampling ratio to the eye region than to other facial parts, thus directing more attention to the eye region.

For training, we utilize the Adam optimizer with a learning rate of $5\times10^{-4}$ and exponential learning decay. The sample ratio is set to 0.8, while the eye region sample ratio is set to 0.5 during training. Each subject is trained for 400K epochs using four A100 GPUs.

\section{Experiments}

\subsection{Setup}
\textbf{Data Preparation.}
We select ETH-XGaze \cite{Zhang_2020_ECCV} for experiments since the dataset contains rich eye movement.
ETH-XGaze is a large-scale gaze estimation dataset collected from 110 subjects with 18 cameras. The dataset provides over one million high-resolution images. We crop the face patch from raw images and resize it into $512\times512$ for training.
Our model is trained based on a sequence of multi-view images.
We select 9 frames which roughly cover the gaze in nine different directions for each subject.
We remove the view under extreme pose and each frame finally provides 13-view images, \textit{i.e.}, we use $9\times13$ images for training on one subject. 

\noindent \textbf{Evaluation Metrics.}
We conduct qualitative and quantitative comparison to demonstrate image quality, where SSIM \cite{wang2004image}, PSNR, and LPIPS \cite{zhang2018unreasonable} are reported for quantitative comparison. 
We also show the advantage of our method in a downstream task.
We render images for data augmentation in gaze estimation task.
We use angular degree error for gaze estimation metric~\cite{Cheng_2021_arxiv}.

\vspace{-0.5mm}
\subsection{Face Rendering under Novel Pose and Gaze}

\begin{figure*}[t]
\centering
\includegraphics[width=1.0\linewidth]{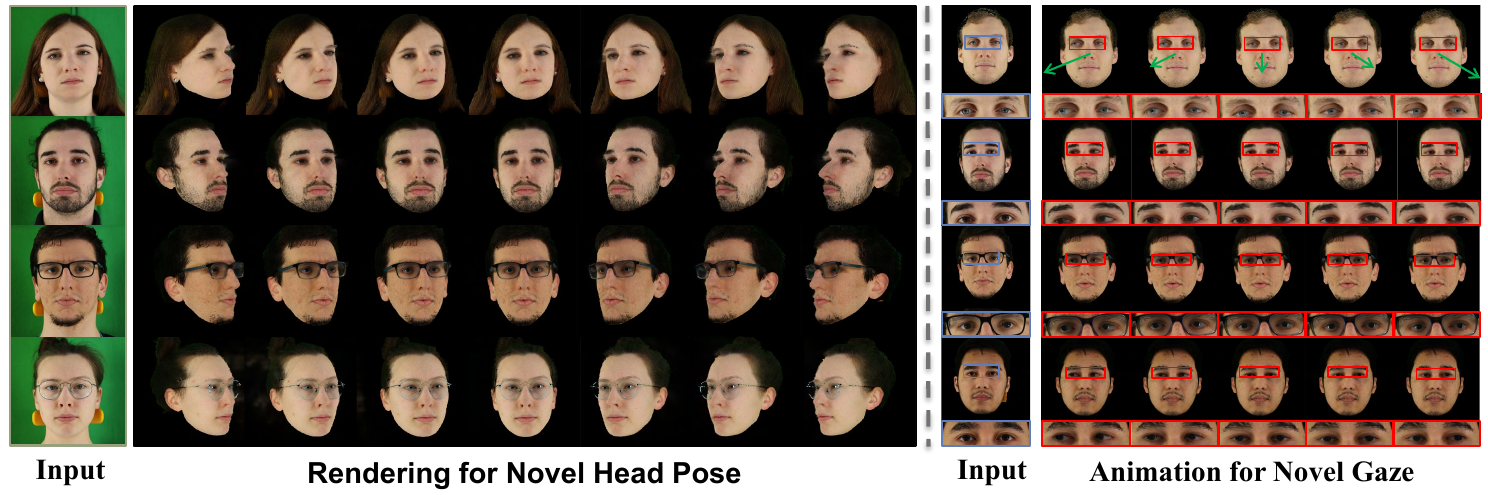}
\vspace{-0.5cm}
\caption{
Face rendering under novel gaze and novel head poses. We train DeNeRF with multi-view images from only nine frames. 
The left images show the rendering under novel head pose. DeNeRF preserves great multi-view consistency.
We also use the DeNeRF to render images under novel gaze in the right image.
DeNeRF is a parametric NeRF model. We can directly change the eyeball pose to generate images under novel gaze. 
DeNeRF generates high-fidelity face images in a large range of head pose and gaze direction. This is the key advantage of DeNeRF.}
\label{fig:novel_pose_gaze}
\end{figure*}

We present the qualitative results for image generation under novel gazes and head poses in Fig. \ref{fig:novel_pose_gaze}. The image on the left showcases the rendered images under novel head poses. Our DeNeRF model excels in generating high-fidelity face images while maintaining multi-view consistency. The skin and eye textures are clearly visible with vivid details, and the hair is accurately reconstructed. Surprisingly, the last two rows in the image demonstrate our model's remarkable generative ability in reconstructing subjects wearing glasses. 
It shows that our DeNeRF can effectively organize the multi-view information from sparse views for high-fidelity 3D face reconstruction.
The image on the right showcases the eye animation generated under novel gaze directions. The animation displays a natural and continuous eye movement despite being trained only on nine sparse gaze directions.
It demonstrates that our model has successfully learned to accurately represent the rotation of the eyeball. This is attributed to the precise eyeball pose achieved through multi-view face tracking using FLAME, as well as the deformation strategy based on canonical space.
They enable us to integrate all information from multi-view images. Overall, the results clearly demonstrate the effectiveness of our model in both face reconstruction and eye animation.

\vspace{-1mm}

\subsection{Comparison with Face Rendering Methods}

We conduct a comprehensive comparison of our method with the SOTA methods in 2D and 3D face rendering with eye animation (STED \cite{zheng2020self} and GazeNeRF \cite{ruzzi2022gazenerf}) and a NeRF-based face rendering approach called HeadNeRF \cite{hong2022headnerf}, which can be adapted for eye animation. 
STED proposes an encoder-decoder structure to automate the disentanglement of gaze direction and head pose. GazeNeRF is a 3D-aware gaze redirection model that takes multi-view images and gaze labels as input and allows control over eye deformation via gaze input. HeadNeRF is a NeRF-based method for high-fidelity facial imaging and can be extended to include gaze direction as additional input for eye animation.
Our comparison primarily focuses on the quality of rendered images, particularly in the eye region.

\begin{wrapfigure}{r}{6.5cm}
\centering
  \footnotesize
  \begin{tabular}{l | c | c | c}
    \toprule
     \textbf{Methods} & \textbf{SSIM~$\uparrow$} & \textbf{PSNR~$\uparrow$}  & \textbf{LPIPS~$\downarrow$} \\
    \midrule
    \textbf{STED~\cite{zheng2020self}} & 0.726 & 17.279 & 0.306 \\
    \textbf{HeadNeRF~\cite{hong2022headnerf}} & 0.718 & 15.262 & 0.300 \\
    \textbf{GazeNeRF~\cite{ruzzi2022gazenerf}} & 0.728 & 15.322 & 0.297 \\
    \midrule
    \textbf{Ours} & \textbf{0.732} & \textbf{19.144} & \textbf{0.265} \\
    \bottomrule
  \end{tabular}
\tabcaption{We show the quantitative comparison between DeNeRF and other SOTA methods in terms of rendered image quality. DeNeRF shows significant improvement in all metrics.}
\label{table:image quailty}
\end{wrapfigure}

\begin{figure}[t]
\centering
\includegraphics[width=0.8\linewidth]{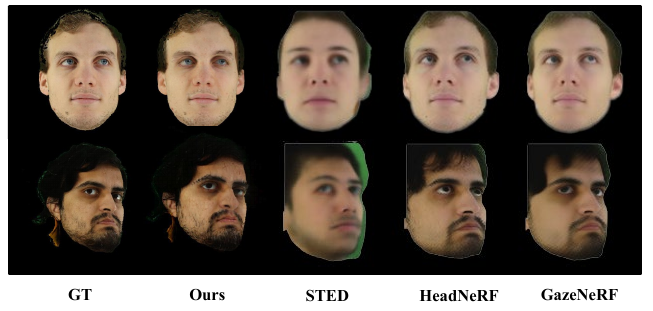}
\vspace{-0.3cm}
  \caption{
  We show the comparison with face rendering methods. We report the result of compared methods from the SOTA face rendering method \cite{ruzzi2022gazenerf}.
  STED and GazeNeRF are designed for gaze redirection. They perform rotation in the feature space which degrades the quality of rendering images. 
  HeadNeRF is adapted for eye animation using gaze direction as additional inputs.
  It is obvious that all compared methods have artifacts in the rendering images,
  while our method renders high-fidelity face images. This demonstrates the advantage of DeNeRF. }
  \label{fig:sota}
\end{figure}

Fig. \ref{fig:sota} shows the qualitative comparison with the SOTA methods. 
All of the methods are capable of face reconstruction. However, the faces generated by STED are blurry. Although the results from HeadNeRF and GazeNeRF are better than STED, they still struggle with achieving a natural skin and hair look. In contrast, our model not only produces a photo-realistic eye region but also accurately reconstructs other facial features and hair.
In the region surrounding the eye, our method produces sharper images than other methods and the finer details of the periocular region, such as eyebrows and eyelids are clearly visible. 

In addition, we show the quantitative results in Table \ref{table:image quailty}.
It shows that our method outperforms GazeNeRF by a large margin which supports the qualitative results above. 
GazeNeRF learns to map the latent code space onto various 3D faces. In contrast, DeNeRF defines a canonical space and warps each point in the observation space to a static canonical space where the deformation is handled by explicit face poses and learned linear blend skinning weights. Such a design provides the neural network with a clear objective of learning the static canonical space, which makes training much easier.

It is worth noting that our model is trained on only 13 views from 9 frames, whereas GazeNeRF is first pre-trained on all available 80 subjects in ETH-XGaze before being fine-tuned using all 18 available views from 100 frames. Despite the vast difference in the amount of training data, our model significantly outperforms GazeNeRF in rendered image quality.

\subsection{Improving Gaze Estimation Performance}
Our method is capable of rendering face images from novel views and eyeball poses. Furthermore, we can generate pseudo-labels for the rendered images based on eyeball rotation and head pose. 
The eyeball rotation is estimated through multi-view face tracking, while the head pose can be derived from camera pose. 
By combining the eyeball rotation and head pose, we can accurately calculate the final gaze direction.

\begin{wrapfigure}{r}{8.3cm}
 \centering
\footnotesize
  \setlength\tabcolsep{4.5pt}
   \begin{tabular}{c | c | c | c | c}
    \toprule
    & \textbf{EyeDiap} \cite{Mora_2014_ETRA} & \textbf{MPIIFace} \cite{Zhang_2017_tpami} & \textbf{RT-Gene} \cite{Fischer_2018_ECCV} & \textbf{Gaze360} \cite{Kellnhofer_2019_ICCV}\\
    \midrule
    \textcolor{red}\texttimes & 41.829 & 36.441 & 42.172 & 17.709  \\
   ${{\color[RGB]{0,170,80}\surd}}$ & \textbf{34.074} & \textbf{28.567} & \textbf{34.632} & \textbf{16.351} \\
    \bottomrule
  \end{tabular}
  \tabcaption{We use rendering images to enhance gaze estimation performance. We train GazeTR in the four datasets and test it in the ETH-XGaze. The {\color[RGB]{0,170,80}checkmark} means we add rendering images into training set. }
  \label{table:gaze error}
\end{wrapfigure}

To demonstrate the capacity of our model in enhancing gaze estimation, we conducted experiments on four renowned gaze datasets including EyeDiap \cite{Mora_2014_ETRA}, MPIIFace \cite{Zhang_2017_CVPRW}, RT-Gene \cite{Fischer_2018_ECCV}, and Gaze360 \cite{Kellnhofer_2019_ICCV}. 
Due to the scarcity of annotated images per subject in the test person specific set of ETH-XGaze, we randomly select seven subjects from the train set of ETH-XGaze for the experiment. We train our model on nine frames for each subject and then generate 324 images under random 36 head poses and nine gaze directions.
These newly generated data were incorporated into the four gaze datasets to create their corresponding augmented version. Finally, we evaluated the performance of the augmented versions on the annotated data of each subject.
We trained the state-of-the-art gaze estimator, GazeTR \cite{cheng2022icpr}, separately on each of the four original datasets and their augmented versions for comparison. 

We present the gaze error in Table \ref{table:gaze error}. The rendering images bring significant improvements on the EyeDiap, MPIIFace, and RT-Gene datasets, with gains of 18.54\%, 21.61\%, and 17.88\%, respectively. This improvement can be attributed to the ability of DeNeRF to generate a wider range of gaze and head poses than those present in the narrow dataset. This increased variation allows for more accurate gaze and head pose estimations.
Despite Gaze360 already having a large gaze and head pose range, our augmented dataset still outperforms it with a 7.67\% improvement. This result further highlights the potential of our method for the gaze estimation task.

\subsection{Ablation Studies}

\begin{figure*}[t]
\centering
\includegraphics[width=\linewidth]{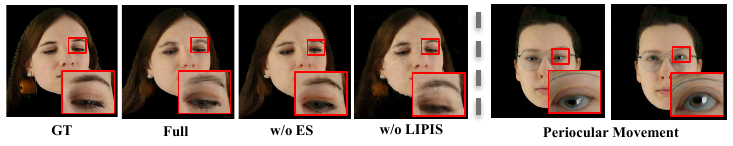}
\vspace{-0.5cm}
\caption{
Ablation study on eye mask guided sampling (ES) and LPIPS loss. The result shows both ES and LPIPS loss improve the image quality in eye region. We design the non-rigid transformation part in the eye deformation field. The right figure demonstrates the effectiveness of the non-rigid transformation. It is obvious that our model accurately capture the periocular movement. This proves the advantage of the eye deformation field.}
\label{fig:ablation}
\end{figure*}

We conduct ablation studies on eye mask guided sampling and LPIPS loss, as seen in Fig. \ref{fig:ablation}. Our model without eye mask guided sampling generates an unnatural eyeball that is nearly completely black. The iris, which is the most crucial semantic information for gaze, is difficult to identify. In contrast, our full model produces a photo-realistic eyeball with a clear boundary between the iris and sclera.
The effectiveness of our eye mask guided sampling can be attributed to its ability to address the deformation issue of small regions. Additionally, when our model is not trained with LPIPS loss, the rendered image appears blurry not only in the eye region but also in other face parts. In comparison, our full model produces sharper details in the rendered image, emphasizing the importance of the LPIPS loss for the image quality of our model.
On the right image, we display the periocular movement as the subject look from bottom to top. It demonstrates that our model is capable of handling non-rigid periocular deformation.

\section{Conclusion}

In this paper, we present a novel dynamic eye-aware NeRF that allows for facial rendering from different perspectives and eye poses. DeNeRF utilizes multi-view images to train a NeRF model of the face. First, we perform face tracking on the multi-view images to capture the eye pose. Then, we fit a parametric 3D face model, FLAME, considering the multi-view consistency. Next, we construct a rotation-aware manifold to model the rotation of the eyeball. We define a canonical space for DeNeRF and transform 3D points from different observation spaces into this space. Finally, we learn a unified face NeRF model on the canonical space while considering an eye deformation field for the transformation. The eye deformation field accounts for rigid transformation, including eyeball rotation, and non-rigid transformation, such as periocular deformation. We evaluate our method on the ETH-XGaze dataset and show that it can render high-fidelity face images from novel viewpoints and eye poses. We also mix our rendered images with the original training set for data augmentation, which further improves performance. In future work, we aim to reduce the requirement for multi-view images and lower computational costs.

\bibliography{egbib}

\end{document}